\documentclass{article}

\usepackage{arxiv}

\usepackage[utf8]{inputenc} 
\usepackage[T1]{fontenc}    
\usepackage[bookmarks=false]{hyperref}       
\usepackage{url}            
\usepackage{booktabs}       
\usepackage{amsfonts}       
\usepackage{nicefrac}       
\usepackage{microtype}      
\usepackage{lipsum}		
\usepackage{graphicx}
\usepackage{natbib}
\usepackage{doi}
\usepackage{xcolor}
\usepackage{graphicx}
\usepackage{booktabs}
\usepackage{multirow}
\usepackage{cleveref}
\usepackage{xcolor}
\usepackage{caption}
\usepackage{float}
\usepackage{subcaption}

\title{Enhancing people localisation in drone imagery for better crowd management by utilising every pixel in high-resolution images}

\date{} 					

\author{ \href{https://orcid.org/0000-0003-1601-6560}{\includegraphics[scale=0.06]{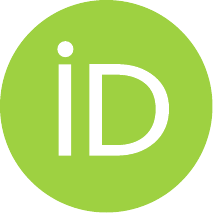}\hspace{1mm}Bartosz Ptak} \\
	Institute of Robotics and Machine Intelligence\\
        Poznań University of Technology\\
	Piotrowo 3A, Poznań, 60-965, Poland \\
	\texttt{bartosz.ptak@doctorate.put.poznan.pl} \\
	\And
	\href{https://orcid.org/0000-0001-6483-2357}{\includegraphics[scale=0.06]{orcid.pdf}\hspace{1mm}Marek Kraft}\thanks{Corresponding author} \\
	Institute of Robotics and Machine Intelligence\\
        Poznań University of Technology\\
	Piotrowo 3A, Poznań, 60-965, Poland \\
	\texttt{marek.kraft@put.poznan.pl}  \\
}

\renewcommand{\shorttitle}{Preprint submitted to Engineering Applications of Artificial Intelligence}

\hypersetup{
pdftitle={A template for the arxiv style},
pdfsubject={q-bio.NC, q-bio.QM},
pdfauthor={David S.~Hippocampus, Elias D.~Striatum},
pdfkeywords={First keyword, Second keyword, More},
}

\begin{document}
\maketitle

\begin{abstract}
Accurate people localisation using drones is crucial for effective crowd management, not only during massive events and public gatherings but also for monitoring daily urban crowd flow. Traditional methods for tiny object localisation using high-resolution drone imagery often face limitations in precision and efficiency, primarily due to constraints in image scaling and sliding window techniques. To address these challenges, a novel approach dedicated to point-oriented object localisation is proposed. Along with this approach, the Pixel Distill module is introduced to enhance the processing of high-definition images by extracting spatial information from individual pixels at once. Additionally, a new dataset named UP-COUNT, tailored to contemporary drone applications, is shared. It addresses a wide range of challenges in drone imagery, such as simultaneous camera and object movement during the image acquisition process, pushing forward the capabilities of crowd management applications. A comprehensive evaluation of the proposed method on the proposed dataset and the commonly used DroneCrowd dataset demonstrates the superiority of our approach over existing methods and highlights its efficacy in drone-based crowd object localisation tasks. These improvements markedly increase the algorithm's applicability to operate in real-world scenarios, enabling more reliable localisation and counting of individuals in dynamic environments.
\end{abstract}


\section{Introduction}\label{sec:Introduction}

The recent decrease in prices and enhancements in the user experience of Unmanned Aerial Vehicles (UAVs) has led to a surge in popularity. However, it has also presented new challenges. From the perspective of urban population growth, the application of UAVs in urban and societal sensing is equally significant. Researchers are increasingly focusing on issues such as road safety management and traffic control~(\cite{outay2020applications, dronova2022unmanned, butilua2022urban}), as well as the utilisation of drones to provide reliable services in Smart Cities~(\cite{mohamed2020unmanned, kharchenko2022uav}). The vantage point provided by aerial imagery enables capturing images of large areas. Still, it brings forth additional challenges arising from the scale variation of the objects of interest, leading to the development of dedicated algorithms~(\cite{li2020object,sun2022fair1m}). One area that has seen significant advancements is the localisation of extremely small and dense objects, such as individual persons in a crowd, with the use of low-altitude aerial imagery. Usually, those objects belong to the micro (less than 2 pixels) or very tiny (diameter of 2-8 pixels) categories, depending on the image's resolution~(\cite{kos2023deep}). Object localisation, which primarily involves estimating the centre coordinates of objects within an image, is particularly advantageous when precise boundary detection is not essential for the intended analysis. The presented approach focuses on identifying the presence and localisation of objects within the scene.
Moreover, creating point-oriented datasets is less costly and time-consuming compared to annotating bounding boxes for object detection. Monitoring moving objects in crowded areas using UAV imagery not only facilitates crowd analysis during large public events~(\cite{husman2021unmanned, ghamari2022unmanned, xu2023assistance}) but also serves as a critical task for safety and surveillance purposes~(\cite{motlagh2017uav}). Compared to fixed-position camera systems~(\cite{hong2023development}), drone-based systems encounter other challenges, requiring the development of algorithms that address issues such as scale variation caused by altitude and perspective changes and camera movement distortions~(\cite{tang2023survey}). With technological advancements, drones are now equipped with high-resolution cameras capable of capturing detailed information about the observed scene~(\cite{bakhtiarnia2022efficient}). However, traditional approaches to drone crowd object localisation often involve downscaling images or employing sliding window techniques to manage the computational complexity of processing large scenes. These methods have inherent limitations that compromise both accuracy and efficiency. Scaling down images may result in information loss, particularly in scenarios with dense crowds or when fine-grained details are crucial~(\cite{ptak2022board}). Similarly, sliding window approaches suffer from inefficiencies due to redundant computations and additional aggregation steps, potentially losing global context~(\cite{divvala2009empirical,zhang2023oriented}).

\begin{figure}[ht!]
\centering
\includegraphics[width=0.8\columnwidth]{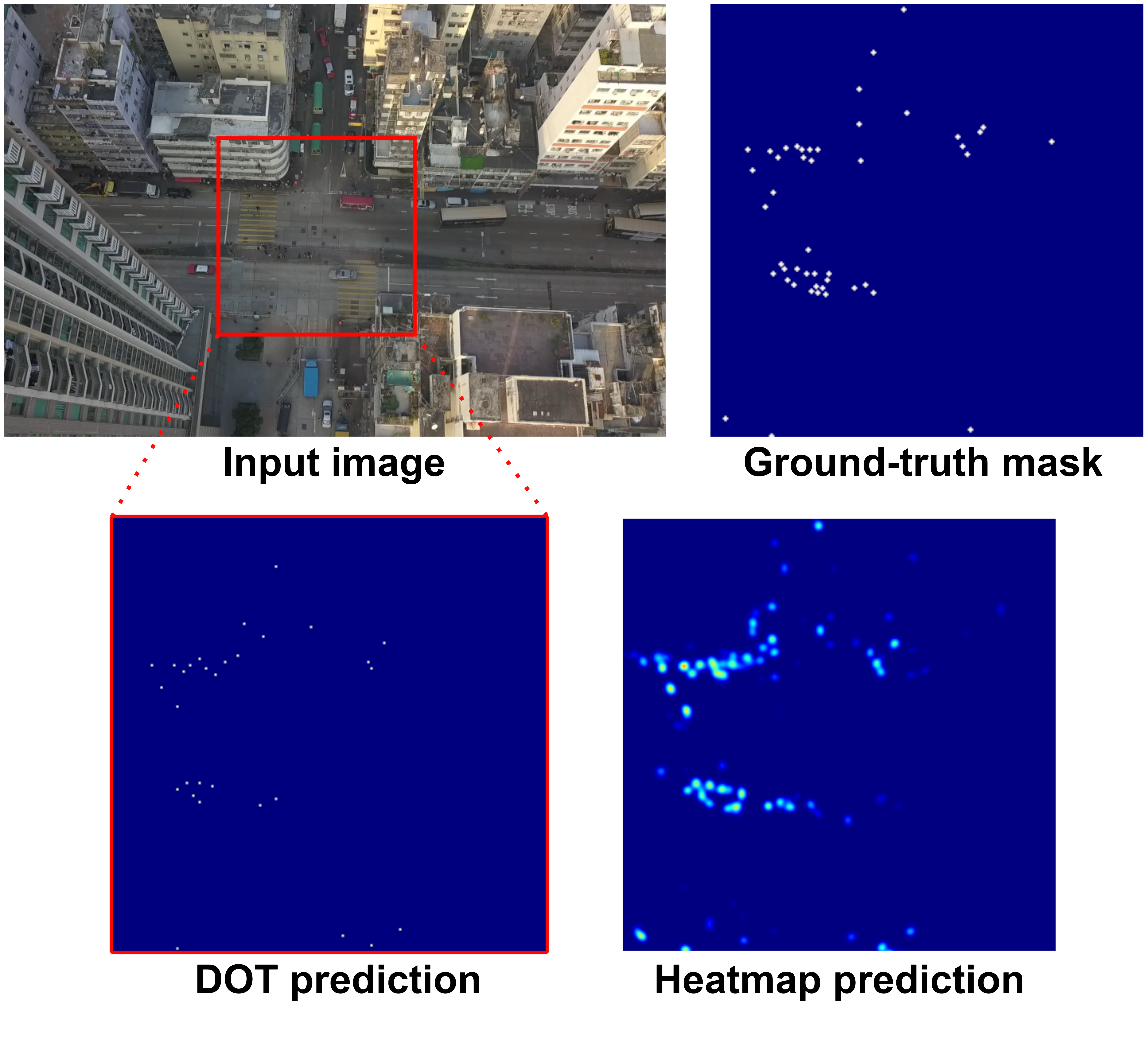}
\caption{A challenging example of tiny people localisation from UAV footage. While existing state-of-the-art methods use Gaussian heatmaps to generate predictions, our Dot approach determines precise people locations.}
\label{fig:abstract}
\end{figure}

To deal with these challenges, we propose a novel approach for object localisation in the UAV-captured imagery, called Dot localisation. This approach is designed to detect point-labelled and dense-invariant objects in images captured by UAVs, defined as a pair of coordinates $(x, y)$ representing the centre of the detected object. By applying a combination of modified focal and point regression losses (see~\Cref{sec:loss}), the method generates more precise masks (\Cref{fig:abstract}). Furthermore, the proposed Pixel Distill (PD) module (\Cref{sec:pd}) enhances the processing of high-definition images, taking into account every pixel value instead of applying a sliding-window approach or interpolation algorithms to reduce the resolution. Moreover, the ablation studies indicate that each proposed element of technical contributions improves model performance. Additionally, by operating on all pixels simultaneously, processing high-resolution images with the proposed PD module is more computationally efficient than processing the image in a window-by-window manner.

Aside from the object localisation method, a novel point-oriented dataset named UP-COUNT (\Cref{sec:upcount}) is released to satisfy the needs of modern drone applications for people localisation. Unlike its predecessors, our new dataset offers a collection of images captured with a drone moving during recording, providing a dynamic perspective that contrasts with the stationary camera setup of the previous datasets. Additionally, our dataset significantly expands the range of people counts, ranging from minimal to high-volume scenarios, and includes altitude information for each image, offering a more detailed and multifaceted dataset for research and analysis. Finally, our method outperforms state-of-the-art methods for both the DroneCrowd dataset~(\cite{wen2021detection}) and the newly introduced dataset. 

The principal contributions of this work are as follows\footnote{The code repository and dataset are available: \url{http://UP-COUNT.github.io/}.}:
\begin{itemize}
    \item We introduce a Dot localisation that outperforms current approaches for the drone crowd localisation task on the DroneCrowd dataset. We also compare our method with various state-of-the-art algorithms from similar remote sensing tasks. 
    \item We propose the Pixel Distill (PD) module, designed to improve the processing of high-resolution images by exploiting the information from every pixel of the scene at once.
    \item We introduce a new benchmark for people localisation in UAV-recorded images called UP-COUNT, simultaneously providing the comparison of the latest methods.
    \item We utilised the proposed method in a real-world scenario, demonstrating the potential of the technology and highlighting directions for further development.
\end{itemize}

\section{Related work}\label{sec:Related_work}

\subsection{General Crowd Object Localisation}

In recent years, a range of methods have been developed for addressing small object localisation through an object counting task, where most of them were related to people. These approaches produce a heatmap mask as their output, which can subsequently be subjected to certain thresholding to achieve object localisation~(\cite{liu2019context,li2018csrnet,wang2020distribution}). More recent methods have shifted the focus to crowd counting through object localisation. Authors of~(\cite{song2021rethinking}) proposed a purely point-based framework for joint counting and individual localisation in crowds, employing the matching of a large set of predefined proposals. A similar but more direct approach was presented in~(\cite{liang2022end}), in which the End-to-End transformer model utilises a direct regression approach and produces point coordinates and confidence scores. A different approach, called Crowd Hat, was proposed in~(\cite{wu2023boosting}). The authors use the advantages of feature maps generated by detection-based methods and propose a plug-and-play module that takes the features and generates a response. The most recent approach is STEERER~(\cite{han2023steerer}), a state-of-the-art method for many ground-view object counting and localisation benchmarks. It addresses scale variations caused by peoples' relative positions to the camera by utilising selectively discriminative features from different scales, adopting feature selection, and segregating discriminative and undiscriminative components of lower-scale features. Regrettably, these methods mainly focus on developing a mechanism to deal with this kind of scale variation, which is not present in top-down UAV imagery. Furthermore, object counting, rather than locating objects, is a prior task in these applications.

\subsection{UAV-based Crowd Object Localisation}

Subsequently, with the release of the DroneCrowd dataset~(\cite{wen2021detection}), the initial approaches to detect crowds from drone imagery emerged. The authors of STNNet~(\cite{wen2021detection}) introduced an approach aimed at addressing density map estimation and object localisation within densely populated scenes captured by drones. In their method, the localisation subnetwork encompasses both classification and regression components. For object localisation, they distribute object proposals across pixels. The classification branch predicts the likelihood that a proposal represents an object, while the regression branch focuses on determining the positions of the positive proposals. In MFA~(\cite{asanomi2023multi}), an advanced technique for crowd localisation is introduced, representing the state-of-the-art in this domain. The authors explore two distinct methods for generating feature maps. The first approach involves heatmap generation, which employs UNet~(\cite{unet}) to estimate object positions' heatmap; each peak in this heatmap signifies the object's location. The second method, Motion and Position Map (MPM) encapsulates position and movement direction through sequential frames and behavioural patterns within the sequence data. These methods employ a sliding-window approach, which can result in the loss of the image's global context and lead to lower performance while also increasing the time needed for processing a single image.

\subsection{Aerial Tiny Object Detection}

Numerous tiny object detection methods dedicated to remote sensing tasks have been proposed, especially after the release of the AI-TOD dataset~(\cite{wang2021tiny}) composed of tiled aerial images with the resolution of $800\times800$ pixels containing tiny objects. The authors of~(\cite{wang2110normalized}) introduced the Normalized Wasserstein Distance (NWD) as a novel metric for bounding box similarity, replacing the conventional Intersection over Union (IoU) and significantly enhancing detection results. Subsequently, the Gaussian Receptive Field-based Label Assignment (RFLA)~(\cite{xu2022rfla}) was introduced to achieve better assignment of object bounding boxes. By leveraging Kullback-Leibler Divergence (KLD) and treating bounding boxes as 2D Gaussian distributions, the authors demonstrated improved scale generalisation of objects compared to NWD. Lastly, Swin-Deformable DEtection TRansformer (SD DETR)~(\cite{liao2023transformer}) was introduced to address very tiny (less than four pixels) objects, becoming state-of-the-art on the AI-TOD dataset. Although these methodologies were originally developed for object detection, they can be adapted for localisation tasks.

\subsection{Focal Loss in Object Localisation}
In this section, we also direct the focus on crowd-localisation approaches that incorporate the utilisation of Focal Loss~(\cite{lin2017focal}) within their frameworks, given its inclusion as an integral component of the loss function proposed in this paper. The authors of~(\cite{liang2022focal}) introduce a novel approach that involves the Focal Inverse Distance Transform (FIDT) map for label generation, together with the proposal of an I-SSIM loss for local Maxima detection. Compared to conventional density maps, their FIDT maps offer precise depictions of individuals' positions, particularly in dense regions, mitigating issues related to overlapping. Additionally, the application of the Independent SSIM loss addresses adverse effects, including blurring and the degradation of local structural details. The study most closely aligned with our work is outlined in~(\cite{zhong2022mask}). In this research, the authors introduce the Mask Focal Loss for crowd people counting, highlighting its effectiveness in enhancing the network's ability to discern head regions and predict heatmaps. They identify the versatility of this loss as a cohesive framework applicable to classification losses rooted in heat maps and binary feature maps. 

\section{UP-COUNT dataset}\label{sec:upcount}

\begin{table}[ht!]
\caption{Comparison between the UP-COUNT dataset and existing DroneCrowd dataset dedicated to drone crowd localisation.}
\label{tab:datasets}

\centering
\resizebox{\columnwidth}{!}{%
\begin{tabular}{|l|c|c|c|c|c|c|c|c|c|}
\hline
\textbf{Dataset} & \multicolumn{1}{l|}{\textbf{Resolution}} & \multicolumn{1}{l|}{\textbf{Frames}} & \multicolumn{1}{l|}{\textbf{Series}} & \multicolumn{1}{l|}{\textbf{\begin{tabular}[c]{@{}c@{}}Min\\ count\end{tabular}}} & \multicolumn{1}{c|}{\textbf{\begin{tabular}[c]{@{}c@{}}Max\\ count\end{tabular}}} & \multicolumn{1}{l|}{\textbf{\begin{tabular}[c]{@{}c@{}}Mean\\ count\end{tabular}}} & \multicolumn{1}{c|}{\textbf{\begin{tabular}[c]{@{}c@{}}Total\\ count\end{tabular}}} & \multicolumn{1}{l|}{\textbf{\begin{tabular}[c]{@{}c@{}}Moving\\ camera\end{tabular}}}  & \multicolumn{1}{c|}{\textbf{\begin{tabular}[c]{@{}c@{}}Drone\\ altitude\end{tabular}}} \\ \hline
DroneCrowd~(\cite{wen2021detection})       & $1920\times1080$          & 33600                               & 112                                     & 25                                                                                & 455                                                                               & 144.8                                                                              & 4864280                                                                           & NO    & NO                                                                                    \\ \hline
UP-COUNT         & $3840\times2160$          & 10000                               & 202                                     & 0                                                                                 & 1039                                                                              & 35.25                                                                              & 352487                                                                             & \textbf{YES}       & \textbf{YES}                                                                   \\ \hline
\end{tabular}%
}
\end{table}

\begin{figure}[ht!]
\centering
\includegraphics[width=0.7\columnwidth]{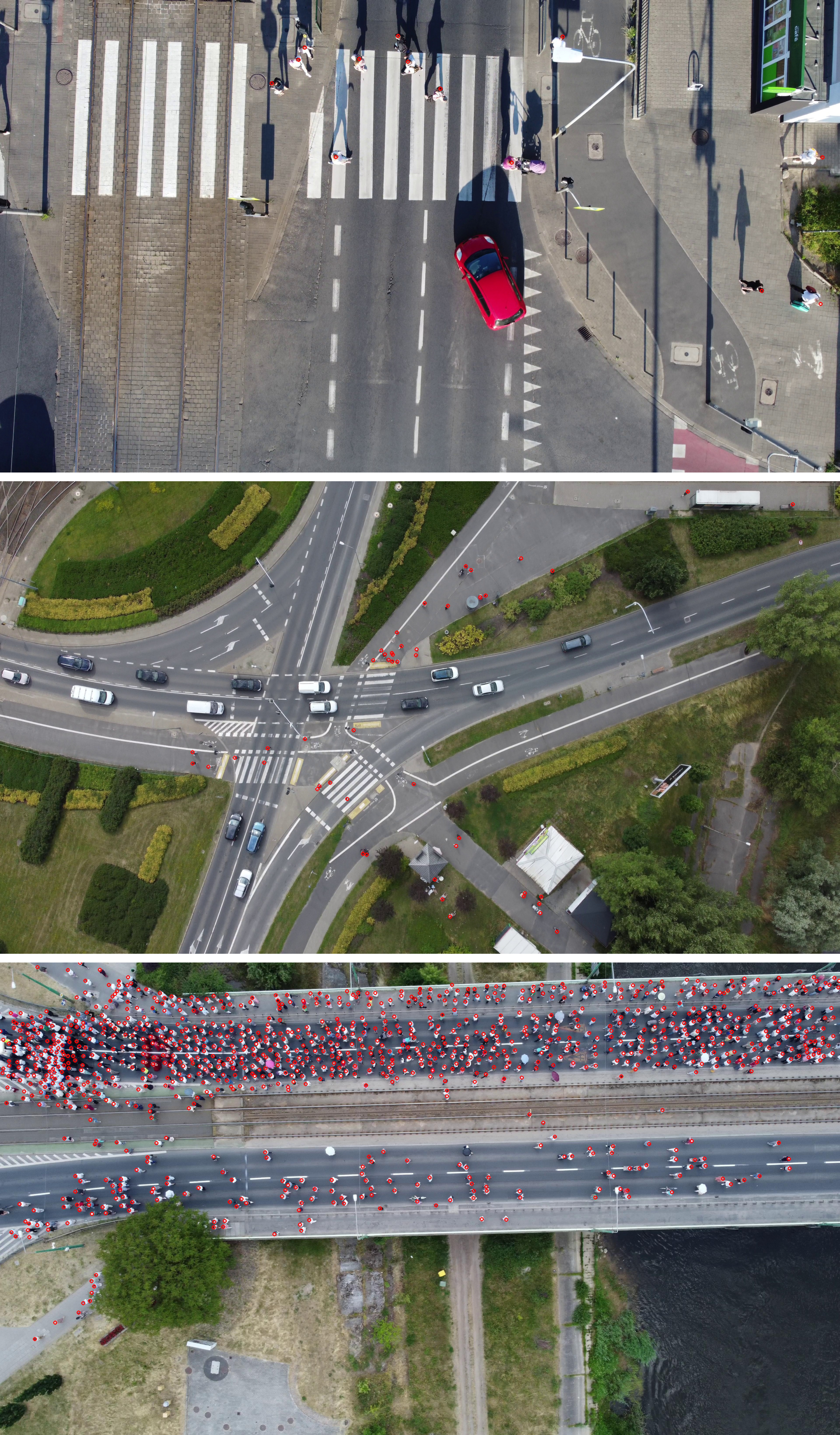}
\caption{Example images with drawn head labels from UP-COUNT dataset. Top - the lowest altitude; middle - the highest; bottom - the maximum crowd image.}
\label{fig:dataset:examples}
\end{figure} 

The newly introduced UP-COUNT dataset includes drone footage captured with cameras from the DJI Mini 2 family UAV. It encompasses diverse environments, including streets, plazas, public transport stops, parks and other green recreation places. We recorded 202 unique videos and then extracted frames with a step of one second, resulting in 10,000 images with a resolution of $3840\times2160$ pixels. The recordings were taken at different altitudes and speeds of flight, and with various densities of people. Acquisition conditions vary in daytime and lighting, creating challenging shadows. Extra altitude information is provided for each image. Next, the labels of people's heads were hand-prepared, resulting in 352,487 instances. During the labelling process, each image was marked and checked by two different people, and the continuity of labels within each sequence was reviewed. \Cref{fig:dataset:examples} contains example frames with the ground truth annotations. The top and middle images represent the lowest- (26.0 meters) and the highest-altitude (101.0 meters) recorded among the sequences, with an average of 60.3 meters. The bottom image presents the most crowded image (1,039 people instances), while the average object count is 35.25. The key features of the dataset are shown in~\Cref{tab:datasets}. Increased variability in crowd counts and different backgrounds caused by the lack of a stationary camera position better reflects real-world scenarios. The UP-COUNT dataset is divided into three subsets for training, validation and testing purposes, containing 141, 30 and 31 sequences, respectively. The described sequences' splits are prepared using altitude-based stratified sampling, providing a comparable altitude distribution between the dataset's splits.

\section{Method}\label{sec:Method}

\begin{figure}[ht!] 
\centering
\includegraphics[width=\columnwidth]{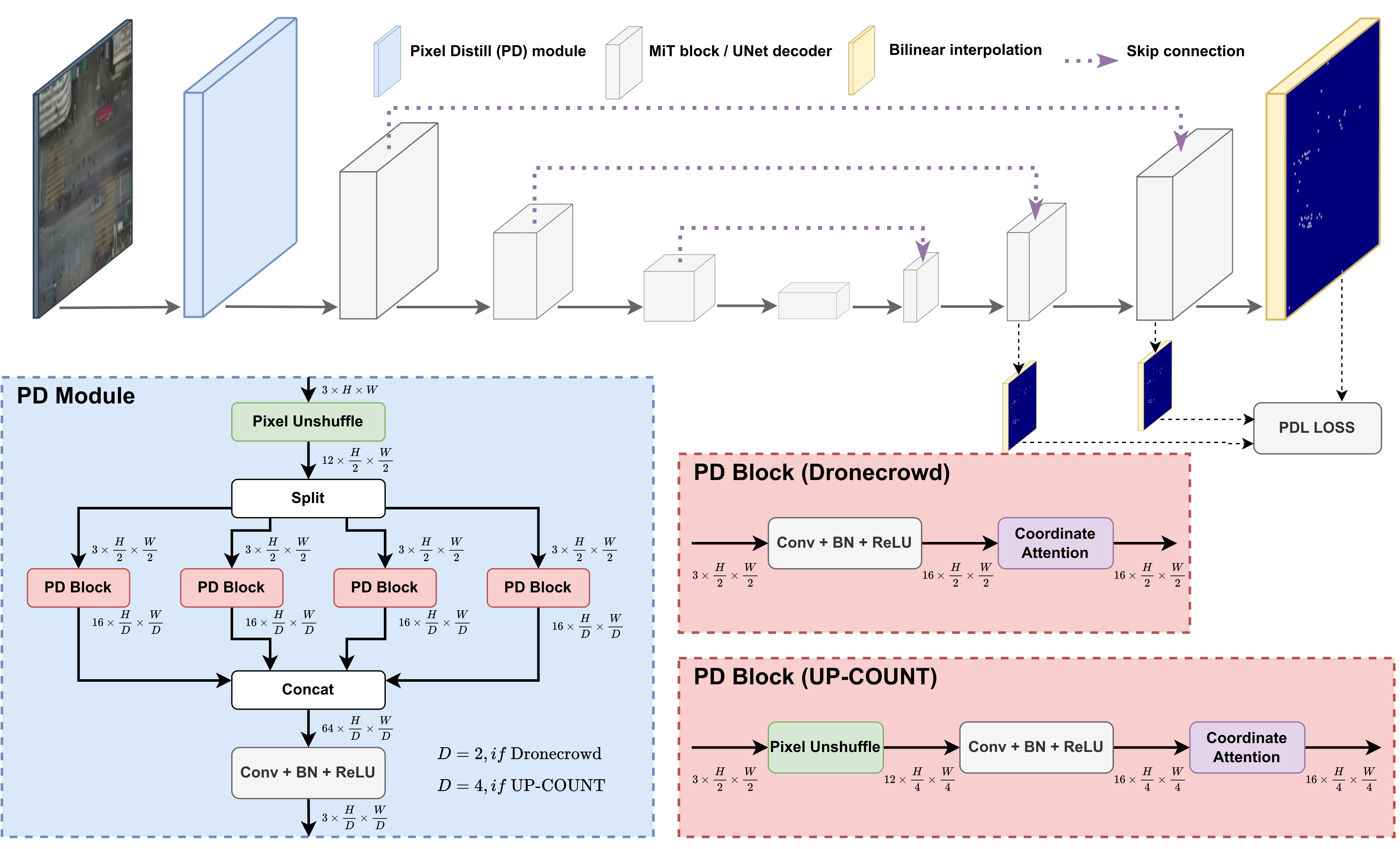}
\caption{Architecture of the designed Dot localisation approach. The Pixel Distill (PD) module processes a full-resolution image and extracts information from each pixel, downsampling by two (for DroneCrowd) or four (for UP-COUNT) times, depending on the PD block used. Since UP-COUNT's resolution is double, the PD block requires an additional resolution reduction.}
\label{fig:pdca}
\end{figure}

The outline of our Dot localisation approach is as follows. \emph{(I)} The full-resolution image is processed with the Pixel Distill module (\Cref{sec:pd}) to extract spatial features. \emph{(II)} Encoder-decoder architecture (\Cref{sec:arch}) generates downsampled masks. \emph{(III)} Masks are upsampled using bilinear interpolation to obtain pixel-perfect accuracy. \emph{(IV)} Objects coordinates are extracted using mechanism described in~\Cref{sec:extr}. The whole architecture is presented in~\Cref{fig:pdca}. 

\subsection{Pixel Distill Module}\label{sec:pd}

The Pixel Distill (PD) module, as illustrated in \Cref{fig:pdca}, represents a versatile architecture for high-resolution image processing and effectively extracts relevant features while preserving spatial information. It is designed to exploit each image pixel, avoiding the information loss that emerges with downsampling interpolation, especially for tiny object localisation. At the early step of the architecture, the PixelUnshuffle layer is applied to reduce the input resolution in half, increasing the number of channels. The layer was proposed in~(\cite{shi2016real}) for super-resolution networks and enables the use of a smaller filter size to integrate the same information while maintaining a given contextual area, reducing computational and memory complexity. Then, the tensor is divided on the channel axis, creating four downsampled images, enabling the processing of each using the Pixel Distill (PD) block. The motivation for using the four-part division lies in its ability to process separate spatial features, enriching the final representation compared to a single block with more filters. We have defined two variants of the PD block: the first variant is tailored for Full HD datasets like DroneCrowd, whereas the second is optimised for 4K images such as UP-COUNT. The two options were proposed to bring an input image to the same output size despite different resolutions. This is because different resolutions require various levels of spatial reduction, ensuring that features are extracted at an appropriate scale for the target dataset. In the first variant, the sixteen-channel feature maps are extracted from the image using a convolutional operation followed by batch normalisation (BN) and ReLU activation. Next, the positional information is extracted and multiplied with feature maps using the Coordinate Attention mechanism~(\cite{hou2021coordinate}). It encodes feature maps as a pair of direction-aware and position-sensitive attention maps that can be complementarily applied to augment the representations of the objects of interest. The second variant of the PD block, in contrast to the first, is preceded by an additional PixelUnshuffle layer for further resolution reduction before processing. After this separable processing, all outputs are concatenated along the channel axis. Then, the final block, consisting of convolution, BN and ReLU layers, is used for channel reduction required by model architecture. This reduction is set to three channels within the proposed module to benefit from weights pretrained on large datasets~(\cite{huh2016makes,risojevic2021we}).

\subsection{Architecture}\label{sec:arch}

The UNet~(\cite{unet}) architecture is chosen to generate accurate masks at a resolution of $960\times544$. It uses skip connections between the successive stages, enabling better encoding of spatial correlation features~(\cite{he2022swin}). The Mix Vision Transformer (MiT), proposed as the encoder of SegFormer~(\cite{xie2021segformer}), is selected as a backbone due to its feature extraction and representation learning capabilities and ablation study results (see \Cref{sec:ablation}). MiT has a lightweight transformer-based hierarchical feature representation and can generate CNN-like multilevel features, offering architecture compatible with the encoder-decoder model. 

\subsection{Mask Extraction and Conversion to Pixel Coordinates}\label{sec:extr}

The proposed architecture generates masks as a direct output and requires some postprocessing steps to obtain $(x, y)$ prediction coordinates. The coordinate calculation is divided into the following steps: 
\begin{enumerate}
    \item Perform \emph{sigmoid} activation to convert the mask's real values to the range of 0 to 1.
    \item Similar to~(\cite{law2018cornernet}), apply non-maximal suppression on a mask by using a $3 \times 3$ max pooling layer to highlight peaks.
    \item Perform mask thresholding to generate a boolean mask where the mask's values are higher than a threshold (in our experiments, $thresh=0.2$).
    \item Select non-zero coordinates.
\end{enumerate}

\subsection{Combined Loss Function}\label{sec:loss}
We propose the loss function named Point Distance-aware Localisation (PDL). The loss is dedicated to drone tiny object localisation with pixel-level precision, generating input image-size masks where 0 represents the background and 1 indicates the object position. It is defined as a combination of three factors: 
\begin{equation}
    L_{total} = 0.25 * L_{neg} + L_{obj} + 2 * L_{reg}
\end{equation}
\noindent where $L_{neg}$ is the modified Focal loss; $L_{reg}$ and $L_{obj}$ are respectively objectness and regression losses described below. The constant coefficients were selected empirically.

\textbf{Modified Focal Loss:} For the ground-truth mask, each object localisation is marked as a positive value while the background is negative. Similarly to~(\cite{law2018cornernet}), we adapted a modified version of Focal loss~(\cite{lin2017focal}). For each positive pixel of a mask, the Focal loss produces a Gaussian neighbourhood. In effect, it reduces the loss penalty for a negative prediction when it is close to a positive one. The $L_{neg}$ value is significantly larger in the initial epochs of the training process and rapidly decreases during it, increasing the relevance of other factors. This approach aims to maximise the distance of the values in the mask corresponding to the objects from the background in an early phase, which helps with convergence and stabilises the training process. We observed that using this loss function also tends to reduce the network's overfitting in comparison to using the mean square error (MSE) loss function (check~\Cref{sec:ablation}). 

\textbf{Point Localisation Loss:} Inspired by the object detection loss function described in~(\cite{wang2023yolov7}), which defines the object detection problem as two separate subtasks of object presence and bounding box position, we propose two loss function components for point objectness and localisation, respectively. The factor of objectness ($L_{obj}$) employs the Binary Entropy criterion, comparing the probabilities of the target and the input mask. This serves to amplify the differentiation between the correct value and the background, thereby enhancing the detection confidence. The regression factor ($L_{reg}$) is introduced to minimise the localisation distance between the prediction and the nearest ground truth, employing the Euclidean distance (L2-norm) to measure the pixel distance between the ground truth pair $(x_1, y_1)$ and the estimated pair $(x_2, y_2)$.

The motivation for using these two loss functions was the maximisation of the difference between background space (zeros) and label space (ones). During experiments, it was observed that the use of Modified Focal Loss in the initial training phase helps with convergence because when its factor is significant, it allows the network to start predicting the first points on the mask. When its importance decreases in favour of Point Localisation Loss, the objectivity and regression criterion becomes dominant, resulting in higher scores. The stabilisation term addresses the observation that the lack of modified Focal Loss required many training epochs before Point Localisation Loss would be reduced.

\subsection{Training Process Details}
We provide details of the training process to aid in reproducing the experiments. The experiments are conducted using a machine equipped with an RTX4090 GPU. Images in HD and 4K resolution are padded to $1920\times1088$ and $3840\times2176$, respectively, to ensure they are divisible by 32 for compatibility with the UNet architecture. Subsequently, standardisation to zero mean and unit variance is performed. The maximal training epochs parameter is set to 100, and the early stopping procedure is defined with the patience of 20. We perform augmentations to improve the model's generalisation capabilities using varied images. Using the fact that the camera's gimbal was directed downward, we generated more samples using the flip-and-rotate transformation, and random resizing. Next, colour-level and noise-adding augmentations were applied to deal with sensor imperfection and light reflections, and random gamma correction was used to produce more overexposed and underexposed samples. Additionally, we implement Mosaic augmentation~(\cite{bochkovskiy2020yolov4}) that combines four images to create a new one.We use the \emph{AdamW} optimiser~(\cite{loshchilov2017decoupled}) which works better for visual transformers~(\cite{guo2022attention}). It is wrapped by the \emph{CosineAnnealingLR} scheduler~(\cite{loshchilov2016sgdr}) with an initial learning rate of $3e^{-4}$ and a period equal to the length of the training epoch. All evaluated architectures were pretrained on Imagenet~(\cite{deng2009imagenet}). Additionally, to improve the training process and handle the problem of vanishing gradients, we applied the Deep Supervision approach~(\cite{wang2015training}) -- a technique that provides a practical and efficient solution to capture fine-grained details in pixel-level tasks better. By implementing deep supervision, auxiliary heads are incorporated into various stages of the network's decoder path, with their individual loss components contributing to the ultimate loss function.

We train the previous state-of-the-art methods using procedures and parameters provided by the authors.

\section{Results}\label{sec:Results}

The proposed method is comprehensively evaluated using two UAV-based datasets: the publicly available DroneCrowd dataset~(\cite{wen2021detection}) that contains 112 drone-captured video clips with $1920 \times 1080$ resolution, and a new dataset named UP-COUNT. The evaluation does not include other aerial datasets such as AI-TOD~(\cite{wang2021tiny}), DOTA~(\cite{9560031}), SODA-A~(\cite{cheng2023towards_soda}), and UAVDT~(\cite{du2018unmanned}) due to their variable resolutions (sometimes very low) or lack of very tiny objects~(\cite{kos2023deep}).

\begin{table}[b]
\caption{The evaluation results for DroneCrowd and UP-COUNT dataset. The best, second, and third-best are shown in {\color[HTML]{32CB00}{\textbf{green}}}, \textcolor{red}{\textbf{red}}, and \textcolor{blue}{\textbf{blue}}, respectively. Acronyms SW and PD mean sliding window approach and Pixel Distill module, respectively.}
\label{tab:sota}
\resizebox{\columnwidth}{!}{%
\begin{tabular}{lcccccccc}
                                                        & \multicolumn{1}{l}{}                                       & \multicolumn{1}{l}{}                                       & \multicolumn{1}{l}{}                                       & \multicolumn{1}{l}{}                                       & \multicolumn{1}{l}{}                                       & \multicolumn{1}{l}{}                                       & \multicolumn{1}{l}{}                                       & \multicolumn{1}{l}{}                                       \\ \hline
\multicolumn{1}{|l|}{}                                  & \multicolumn{4}{c|}{\textbf{DroneCrowd}}                                                                                                                                                                                                          & \multicolumn{4}{c|}{\textbf{UP-COUNT}}                                                                                                                                                                                                             \\ \cline{2-9} 
\multicolumn{1}{|l|}{\multirow{-2}{*}{\textbf{Method}}} & \multicolumn{1}{c|}{\textbf{L-mAP}}                        & \multicolumn{1}{c|}{\textbf{L-AP@10}}                       & \multicolumn{1}{c|}{\textbf{L-AP@15}}                       & \multicolumn{1}{c|}{\textbf{L-AP@20}}                       & \multicolumn{1}{c|}{\textbf{L-mAP}}                        & \multicolumn{1}{c|}{\textbf{L-AP@10}}                       & \multicolumn{1}{c|}{\textbf{L-AP@15}}                       & \multicolumn{1}{c|}{\textbf{L-AP@20}}                       \\ \hline\hline
\multicolumn{1}{|l|}{\textbf{STNNet}~(\cite{wen2021detection})}                   & \multicolumn{1}{c|}{40.45}                                 & \multicolumn{1}{c|}{42.75}                                 & \multicolumn{1}{c|}{50.98}                                 & \multicolumn{1}{c|}{55.77}                                 & \multicolumn{1}{c|}{37.20}                                 & \multicolumn{1}{c|}{28.48}                                 & \multicolumn{1}{c|}{50.97}                                 & \multicolumn{1}{c|}{61.60}                                 \\ \hline
\multicolumn{1}{|l|}{\textbf{MFA}~(\cite{asanomi2023multi})}                      & \multicolumn{1}{c|}{43.43}                                 & \multicolumn{1}{c|}{47.14}                                 & \multicolumn{1}{c|}{51.58}                                 & \multicolumn{1}{c|}{54.02}                                 & \multicolumn{1}{c|}{x}                                     & \multicolumn{1}{c|}{x}                                     & \multicolumn{1}{c|}{x}                                     & \multicolumn{1}{c|}{x}                                     \\ \hline\hline
\multicolumn{1}{|l|}{\textbf{STEERER}~(\cite{han2023steerer})}                  & \multicolumn{1}{c|}{38.31}                                 & \multicolumn{1}{c|}{41.96}                                 & \multicolumn{1}{c|}{46.58}                                 & \multicolumn{1}{c|}{49.07}                                 & \multicolumn{1}{c|}{40.20}                                 & \multicolumn{1}{c|}{42.14}                                 & \multicolumn{1}{c|}{50.32}                                 & \multicolumn{1}{c|}{54.30}                                 \\ \hline\hline
\multicolumn{1}{|l|}{\textbf{RFLA}~(\cite{xu2022rfla})}                     & \multicolumn{1}{c|}{32.05}                                 & \multicolumn{1}{c|}{34.41}                                 & \multicolumn{1}{c|}{39.59}                                 & \multicolumn{1}{c|}{42.52}                                 & \multicolumn{1}{c|}{32.41}                                 & \multicolumn{1}{c|}{33.27}                                 & \multicolumn{1}{c|}{42.54}                                 & \multicolumn{1}{c|}{46.38}                                 \\ \hline
\multicolumn{1}{|l|}{\textbf{SD-DETR}~(\cite{liao2023transformer})}                  & \multicolumn{1}{c|}{{\color{blue} \textbf{48.12}}} & \multicolumn{1}{c|}{52.56}                                 & \multicolumn{1}{c|}{{\color{blue} \textbf{57.35}}} & \multicolumn{1}{c|}{{\color{blue} \textbf{60.08}}} & \multicolumn{1}{c|}{{\color{blue} \textbf{57.89}}}                                 & \multicolumn{1}{c|}{{\color{blue} \textbf{63.57}}}                                 & \multicolumn{1}{c|}{{\color[HTML]{FE0000} \textbf{75.76}}} & \multicolumn{1}{c|}{{\color[HTML]{FE0000} \textbf{79.30}}} \\ \hline\hline
\multicolumn{1}{|l|}{\textbf{Dot}}             & \multicolumn{1}{c|}{47.63}                                 & \multicolumn{1}{c|}{{\color{blue} \textbf{53.37}}} & \multicolumn{1}{c|}{56.99}                                 & \multicolumn{1}{c|}{58.86}                                 & \multicolumn{1}{c|}{{\color[HTML]{FE0000} \textbf{60.66}}} & \multicolumn{1}{c|}{{\color[HTML]{FE0000} \textbf{69.07}}} & \multicolumn{1}{c|}{{\color{blue} \textbf{75.45}}}                                 & \multicolumn{1}{c|}{{\color{blue} \textbf{77.26}}}                                 \\ \hline
\multicolumn{1}{|l|}{\textbf{Dot + SW}}             & \multicolumn{1}{c|}{{\color[HTML]{FE0000} \textbf{50.73}}}                                 & \multicolumn{1}{c|}{{\color[HTML]{FE0000} \textbf{55.61}}} & \multicolumn{1}{c|}{{\color[HTML]{FE0000} \textbf{59.23}}}                                 & \multicolumn{1}{c|}{{\color[HTML]{FE0000} \textbf{61.52}}}                                 & \multicolumn{1}{c|}{54.17} & \multicolumn{1}{c|}{58.74} & \multicolumn{1}{c|}{61.67}                                 & \multicolumn{1}{c|}{63.21}                                 \\ \hline
\multicolumn{1}{|l|}{\textbf{Dot + PD}}                      & \multicolumn{1}{c|}{{\color[HTML]{32CB00} \textbf{51.00}}} & \multicolumn{1}{c|}{{\color[HTML]{32CB00} \textbf{57.06}}} & \multicolumn{1}{c|}{{\color[HTML]{32CB00} \textbf{60.45}}} & \multicolumn{1}{c|}{{\color[HTML]{32CB00} \textbf{62.29}}} & \multicolumn{1}{c|}{{\color[HTML]{32CB00} \textbf{66.49}}} & \multicolumn{1}{c|}{{\color[HTML]{32CB00} \textbf{75.46}}} & \multicolumn{1}{c|}{{\color[HTML]{32CB00} \textbf{79.57}}} & \multicolumn{1}{c|}{{\color[HTML]{32CB00} \textbf{81.16}}} \\ \hline
\end{tabular}%
}
\end{table}

\begin{figure}[ht!] 
\centering
\includegraphics[width=0.85\columnwidth]{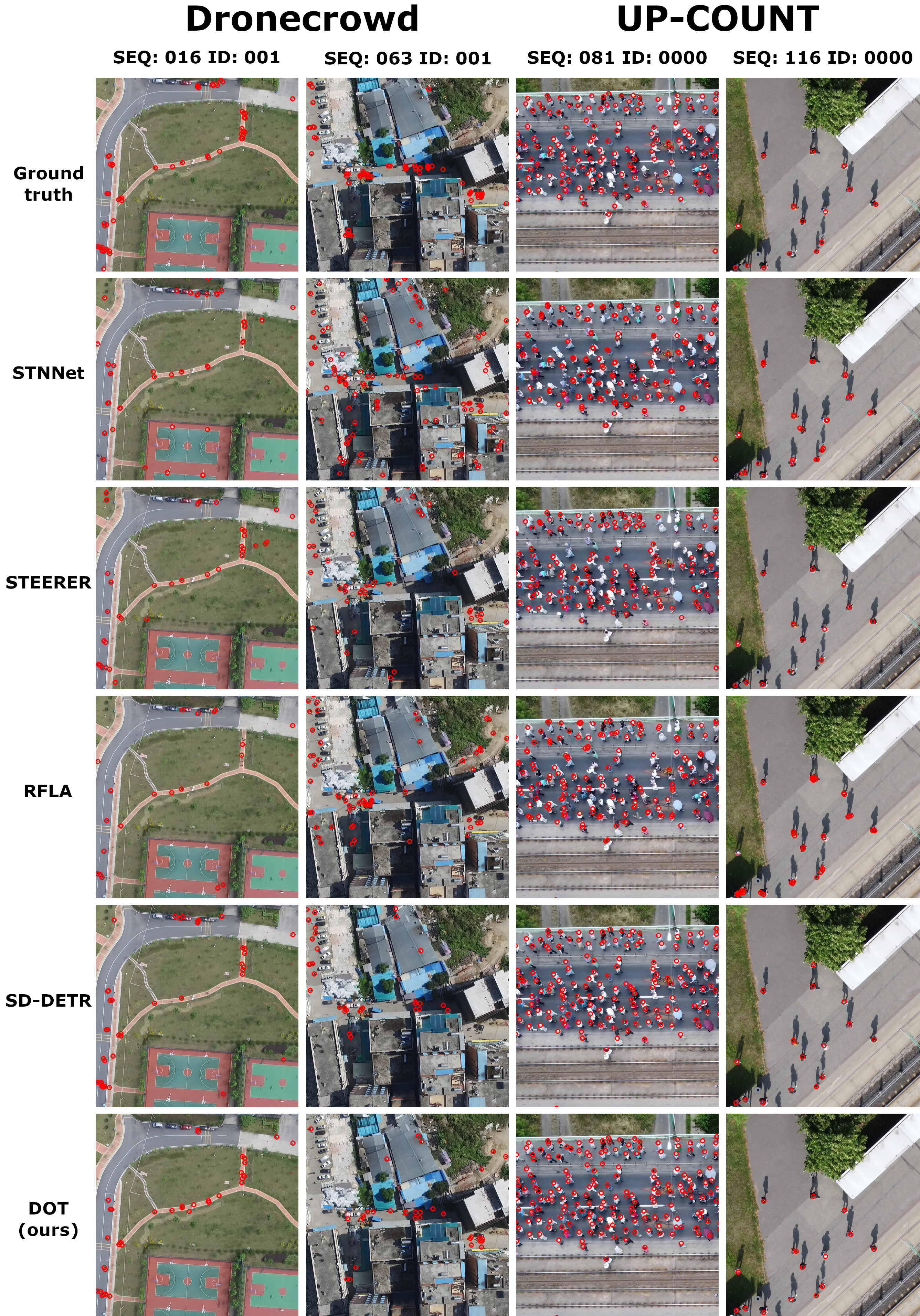}
\caption{Visual comparison of results provided by different methods for both datasets.}
\label{fig:vis_results}
\end{figure} 

The Average Precision for localisation tasks (L-AP) is employed for evaluation. It is determined by the distance threshold calculated using the greedy method, according to the procedure introduced in~(\cite{wen2021detection}). The results are reported for three distance thresholds: 10 pixels (L-AP@10), 15 pixels (L-AP@15), and 20 pixels (L-AP@20), along with L-mAP, which provides cumulative results in thresholds ranging from 1 to 25 pixels.

According to \Cref{tab:sota}, our method has been benchmarked against three groups of methods. Compared to STNNet~(\cite{wen2021detection}) and MFA~(\cite{asanomi2023multi}), previous state-of-the-art methods on the DroneCrowd dataset, our method outperforms them by 10.55 and 7.57 on L-mAP, and 14.31 and 9.92 on L-AP@10, respectively. On the UP-COUNT dataset, STNNet exhibits significantly poorer precision, delivering results that are 29.29 (L-mAP) and 46.98 (L-AP@10) worse. It's important to note that the results for the MFA method are sourced from~(\cite{asanomi2023multi}) as we could not reproduce results on the DroneCrowd dataset and train it using the UP-COUNT dataset. Despite reporting the best results on several non-remote sensing datasets for object localisation, STEERER~(\cite{han2023steerer}) shows low adaptability to the described task, resulting in unsatisfactory performance. Similarly, low metrics were reported by RFLA~(\cite{xu2022rfla}), a method designed specifically for tiny object detection. In contrast, SD-DETR~(\cite{liao2023transformer}), another algorithm from this domain, appears to be more robust. Using the DroneCrowd dataset yields results that are 2.88 and 4.50 worse in terms of L-mAP and L-AP@10, respectively, compared to our method. On the UP-COUNT, the results are lower at 8.60 and 10.89 for the same metrics. The visual comparison of evaluated methods is presented in \Cref{fig:vis_results}.

The final evaluation assesses the robustness achieved through the integration of the Pixel Distill (PD) module within the framework of our method, comparing it against commonly employed techniques: image scaling to match the network input size and utilising a moving window for inference across the entire image to keep original input resolution. Dot localisation employing image bilinear interpolation to decrease image resolution yielded L-mAP scores of 47.63 and 60.66, and L-AP@10 scores of 53.37 and 69.07 for both datasets, respectively. Results obtained from Dot localisation using the sliding window (SW) approach on the DroneCrowd dataset, featuring Full HD images, with images cropped into four parts, yielded L-mAP and L-AP@10 scores of 50.73 and 55.61, respectively. However, notably different outcomes were observed for the second dataset, where 4K images were cropped into sixteen tiles, resulting in a marked decrease in performance. In conclusion, the evaluation exhibits improved performance outcomes of employing the PD module compared to traditional techniques, highlighting their sensitivity to dataset characteristics such as resolution and cropping approach.

In summary, our Dot localisation with the Pixel Distill (PD) module reported state-of-the-art results for DroneCrowd, with L-mAP of 51.00 and L-AP@10 of 57.06. Similarly, it achieved scores of L-mAP 66.49 and L-AP@10 75.46 for UP-COUNT, establishing a baseline for this benchmark.

\subsection{L-mAP's Thresholds Investigation}

\begin{figure}[ht!] 
\centering
\includegraphics[width=0.95\columnwidth]{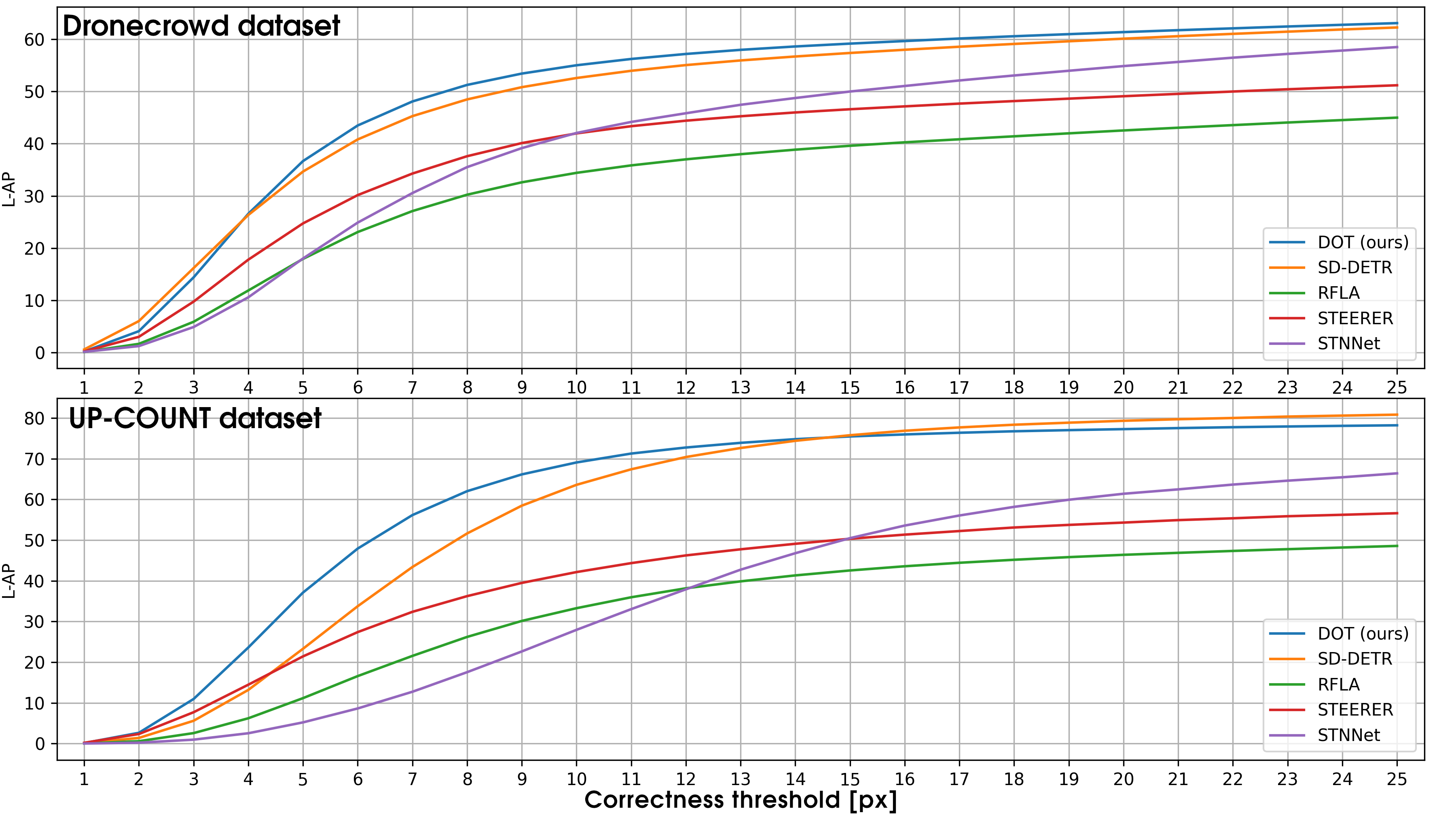}
\caption{L-AP metric comparison regarding the correctness threshold. The range of thresholds covers the L-mAP metric, between 1 and 25 pixels.}
\label{fig:lap_results}
\end{figure}

To provide further insight into the L-mAP metric, the evaluations across a range of thresholds (from 1 to 25) were conducted. Referring to~\Cref{fig:lap_results}, it is worth pointing out that SD-DETR and Dot localisation stand out from the other methods in results. Generally, lower thresholds of correctness imply more precise localisation, while higher thresholds may result in false-positive cases being matched as correct. For DroneCrowd, when the threshold is very low (less than four), the SD-DETR method shows a few better results compared with our method, while the next threshold ranges report better performance of our method. And, for UP-COUNT, our method exhibits significantly higher metrics for thresholds below fifteen, with SD-DETR showing higher performance in higher threshold ranges. This nuanced understanding of threshold-based evaluations provides valuable insights into the comparative strengths of each method under consideration, demonstrating superior performance at lower correctness thresholds. These results demonstrate notable advancements, particularly considering that the Dot model has 12M fewer parameters and provides 3.85x faster inference speed (0.013 vs 0.050 seconds on RTX4090).

\subsection{Ablation Study}\label{sec:ablation}

The ablation study of specific approach components is conducted, with results presented in \Cref{tab:ablation}. This study encompasses three well-known feature extractors: ResNet~(\cite{he2016deep, wightman2021resnet}), which facilitates effective feature representation learning through residual connections, mitigating the vanishing gradient problem; EfficientNet~(\cite{tan2019efficientnet}), which offers the best trade-off between efficiency and performance, capturing intricate patterns from input images while minimising computational resources; MiT B2 described in~\Cref{sec:arch}.

Using the proposed loss function - PDL (\Cref{sec:loss}) yields an average improvement of 10.7 and 10.4 in L-mAP for both datasets compared to the mean square error (MSE) function. Additionally, integrating Pixel Distill results in an average improvement of 3.0 and 2.3 compared to its absence. Employing the MiT B2 encoder leads to higher datasets' L-mAP scores: 51.00 and 65.05, surpassing those of ResNet50 (50.36 and 64.39) and EfficientNetB2 (48.34 and 61.17).

\begin{table}[ht]
\centering
\caption{Proposed Dot model results from evaluation according to the used encoder as a backbone, loss function, and Pixel Distill (PD) module.}
\label{tab:ablation}
\resizebox{0.7\columnwidth}{!}{%
\begin{tabular}{|c|c|c|cc|cc|}
\hline
\multirow{2}{*}{\textbf{Encoder}}        & \multirow{2}{*}{\textbf{Loss}} & \multirow{2}{*}{\textbf{PD}} & \multicolumn{2}{c|}{\textbf{DroneCrowd}}              & \multicolumn{2}{c|}{\textbf{UP-COUNT}}                \\ \cline{4-7} 
                                         &                                &                              & \multicolumn{1}{c|}{\textbf{L-mAP}} & \textbf{mAP@10} & \multicolumn{1}{c|}{\textbf{L-mAP}} & \textbf{mAP@10} \\ \hline\hline
\multirow{3}{*}{\textbf{ResNet50}}       & MSE                            & no                           & \multicolumn{1}{c|}{34.35}          & 38.37           & \multicolumn{1}{c|}{49.46}               &     55.29            \\ \cline{2-7} 
                                         & PDL                            & no                           & \multicolumn{1}{c|}{45.83}          & 50.94           & \multicolumn{1}{c|}{60.66}          & 69.07           \\ \cline{2-7} 
                                         & PDL                            & yes                          & \multicolumn{1}{c|}{50.36}          & 56.04           & \multicolumn{1}{c|}{64.49}          & 73.46           \\ \hline\hline
\multirow{3}{*}{\textbf{EfficientNetB2}} & MSE                            & no                           & \multicolumn{1}{c|}{36.13}          & 40.30           & \multicolumn{1}{c|}{50.98}             & 57.53          \\ \cline{2-7} 
                                         & PDL                            & no                           & \multicolumn{1}{c|}{47.20}          & 52.46           & \multicolumn{1}{c|}{60.34}          & 67.63           \\ \cline{2-7} 
                                         & PDL                            & yes                          & \multicolumn{1}{c|}{48.34}          & 54.08           & \multicolumn{1}{c|}{61.17}          & 68.02           \\ \hline\hline
\multirow{3}{*}{\textbf{MiT B2}}         & MSE                            & no                           & \multicolumn{1}{c|}{38.08}          & 42.51            & \multicolumn{1}{c|}{52.24}         & 58.78            \\ \cline{2-7} 
                                         & PDL                            & no                           & \multicolumn{1}{c|}{47.63}          & 53.37           & \multicolumn{1}{c|}{62.83}          & 70.01           \\ \cline{2-7} 
                                         & PDL                            & yes                          & \multicolumn{1}{c|}{\textbf{51.00}}          & \textbf{57.06}           & \multicolumn{1}{c|}{\textbf{65.05}}          & \textbf{74.03}           \\ \hline
\end{tabular}%
}
\end{table}

\subsection{Cross-dataset Evaluation}

A cross-dataset evaluation was conducted to showcase the Dot approach's generalisation capabilities and the improved diversity of the UP-COUNT dataset. The results, as depicted in \Cref{tab:cross_dataset}, reveal that both methods -- SD-DETR~(\cite{liao2023transformer}) and Dot -- experienced smaller decreases in performance when trained on UP-COUNT and evaluated on DroneCrowd compared to the reverse scenario. Furthermore, Dot outperformed SD-DETR in both instances, achieving higher metrics with L-mAP scores of 15.95 and 24.33 for the DroneCrowd to UP-COUNT and UP-COUNT to DroneCrowd transfers, respectively, compared to SD-DETR's scores of 12.49 and 20.12.

\begin{table}[ht!]
\centering
\caption{Cross dataset evaluation results.}
\label{tab:cross_dataset}
\resizebox{0.6\columnwidth}{!}{%
\begin{tabular}{|c|c|c|c|c|}
\hline
\textbf{\begin{tabular}[c]{@{}c@{}}Training\\ dataset\end{tabular}} & \textbf{\begin{tabular}[c]{@{}c@{}}Testing\\ dataset\end{tabular}} & \textbf{Method} & \textbf{L-mAP} & \textbf{L-AP@10} \\ \hline\hline
\multirow{2}{*}{DroneCrowd}                                         & \multirow{2}{*}{UP-COUNT}                                          & SD-DETR         & 12.49          & 9.89             \\ \cline{3-5} 
                                                                    &                                                                    & Dot             & 15.95          & 14.99            \\ \hline\hline
\multirow{2}{*}{UP-COUNT}                                           & \multirow{2}{*}{DroneCrowd}                                        & SD-DETR         & 20.12          & 21.47            \\ \cline{3-5} 
                                                                    &                                                                    & Dot             & 24.33          & 26.95            \\ \hline
\end{tabular}
}
\end{table}

\section{Real-World Use Case}

\begin{figure}[ht!]
     \centering
     \begin{subfigure}[b]{0.49\columnwidth}
         \centering
         \includegraphics[width=\columnwidth]{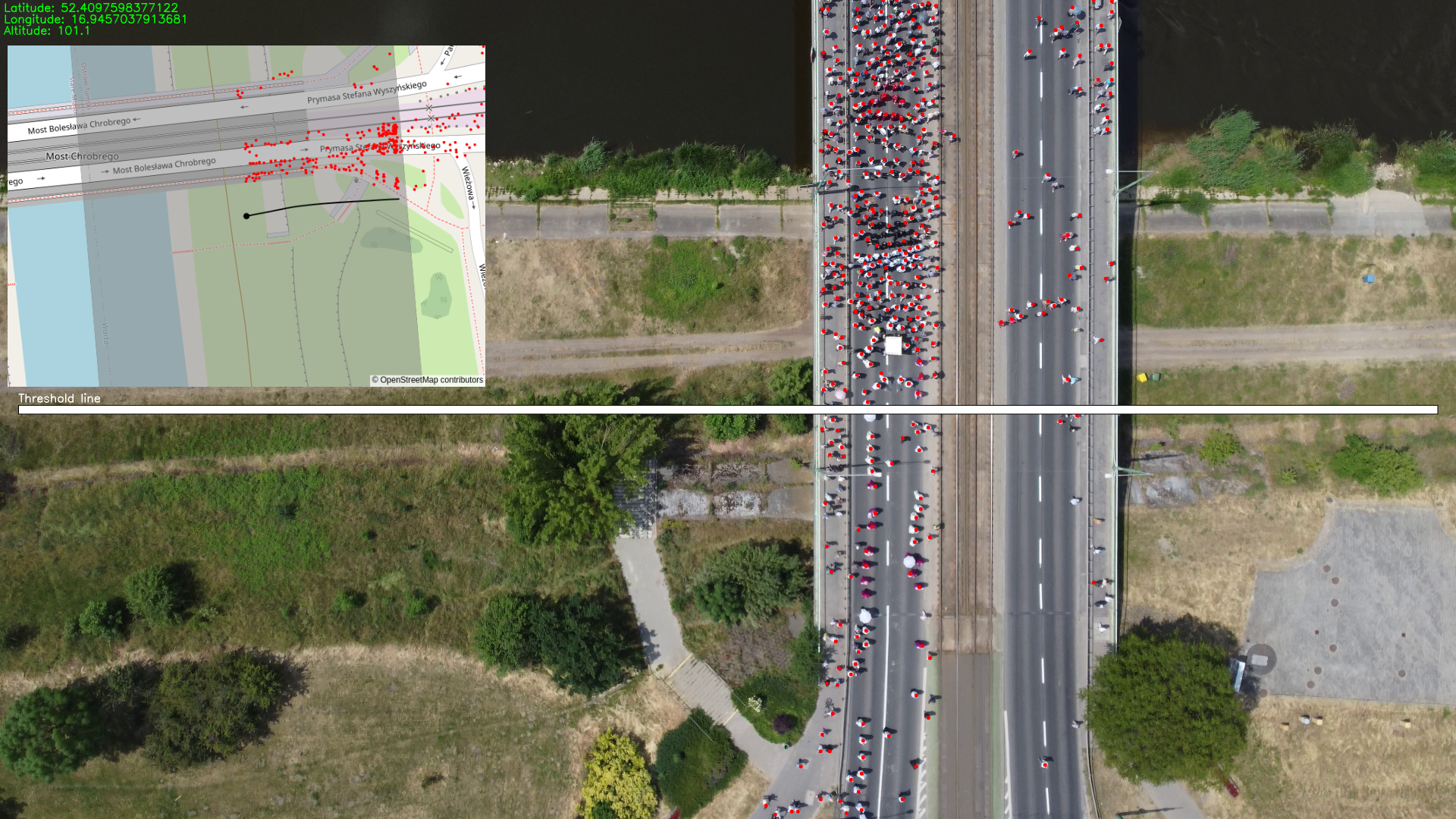}
         \caption{Frame 418}
     \end{subfigure}
     \hfill
     \begin{subfigure}[b]{0.49\columnwidth}
         \centering
         \includegraphics[width=\columnwidth]{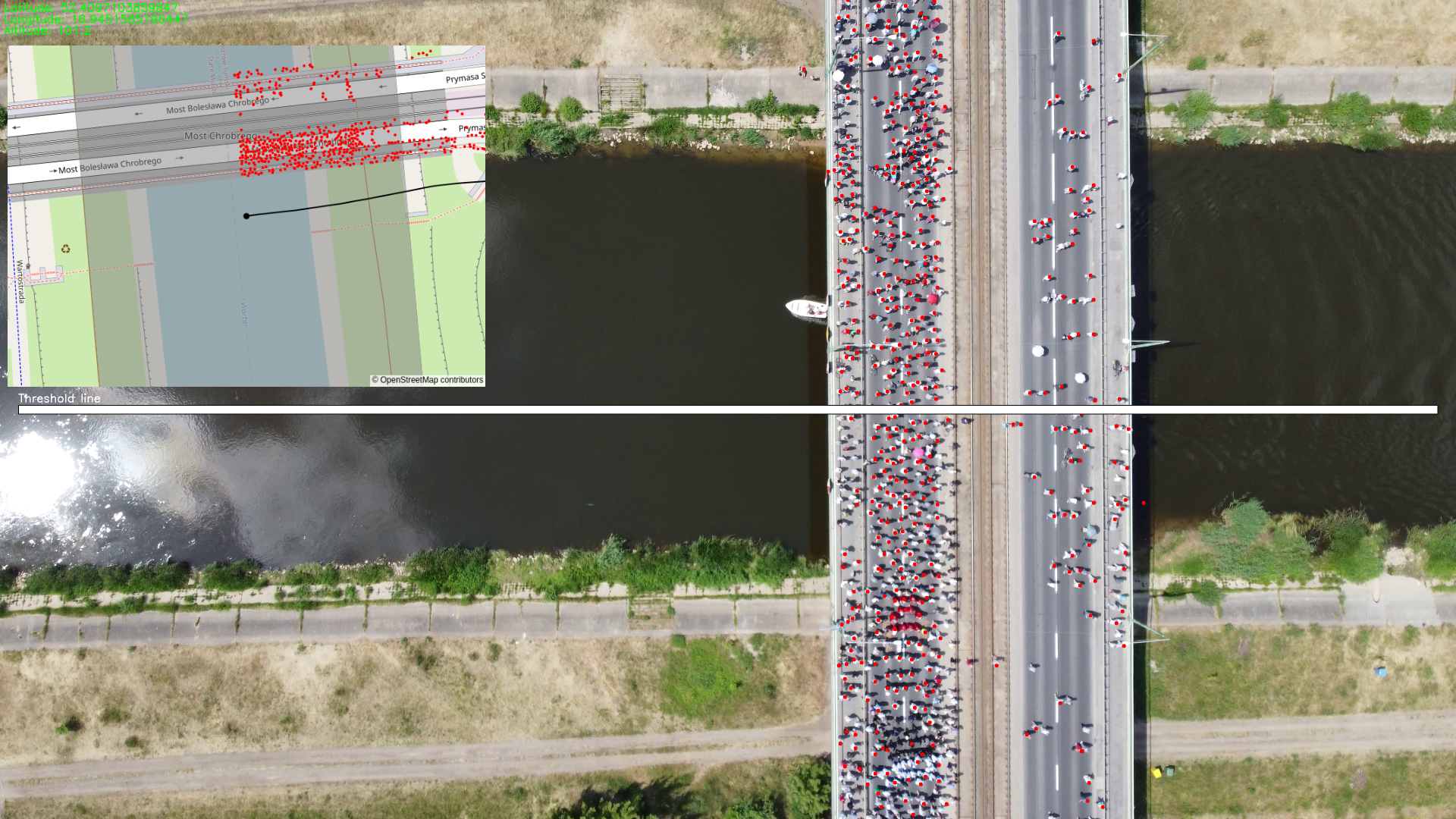}
         \caption{Frame 706}
     \end{subfigure}
     
     \begin{subfigure}[b]{0.49\columnwidth}
         \centering
         \includegraphics[width=\columnwidth]{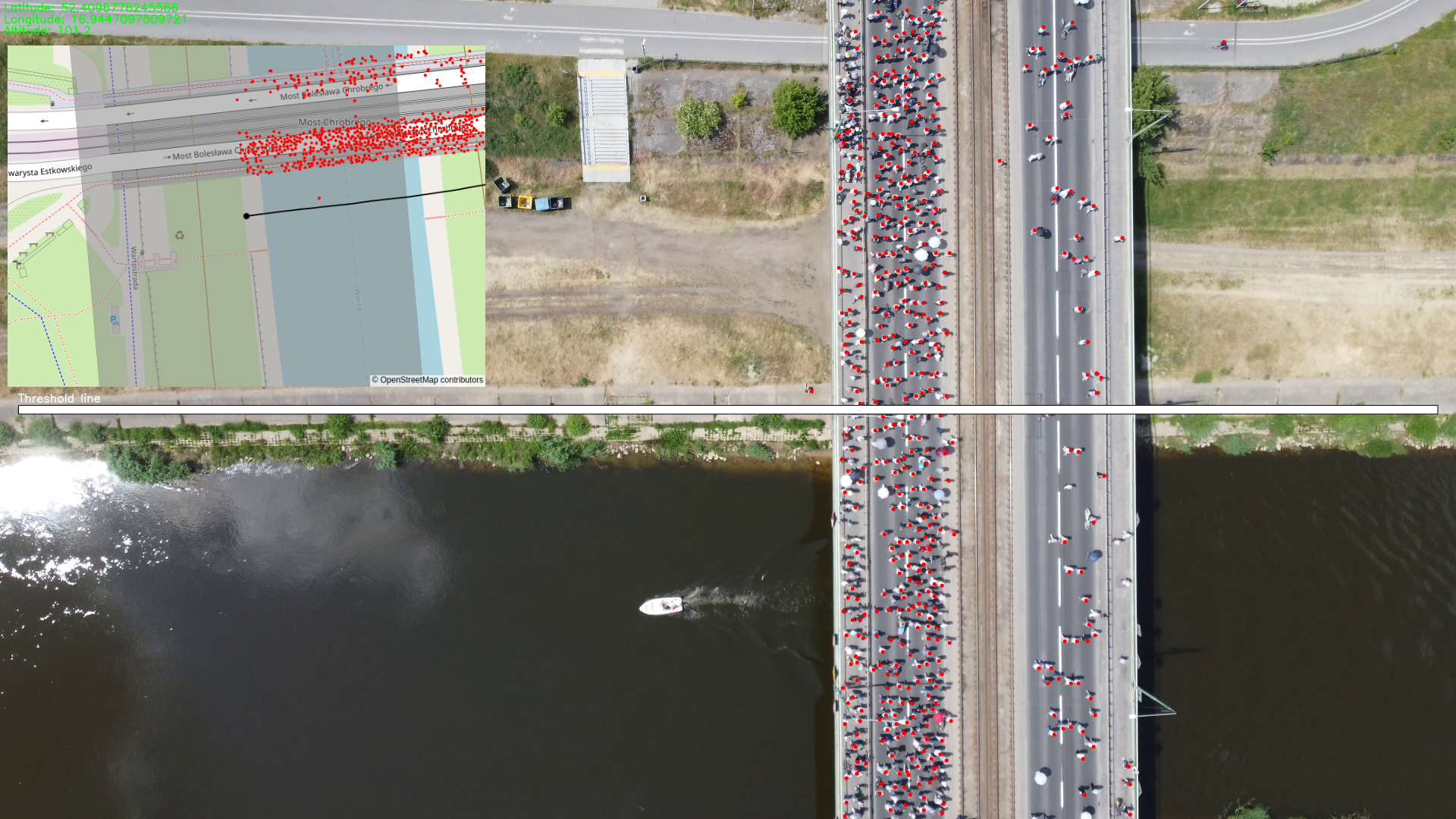}
         \caption{Frame 993}
     \end{subfigure}
    \hfill
     \begin{subfigure}[b]{0.49\columnwidth}
         \centering
         \includegraphics[width=\columnwidth]{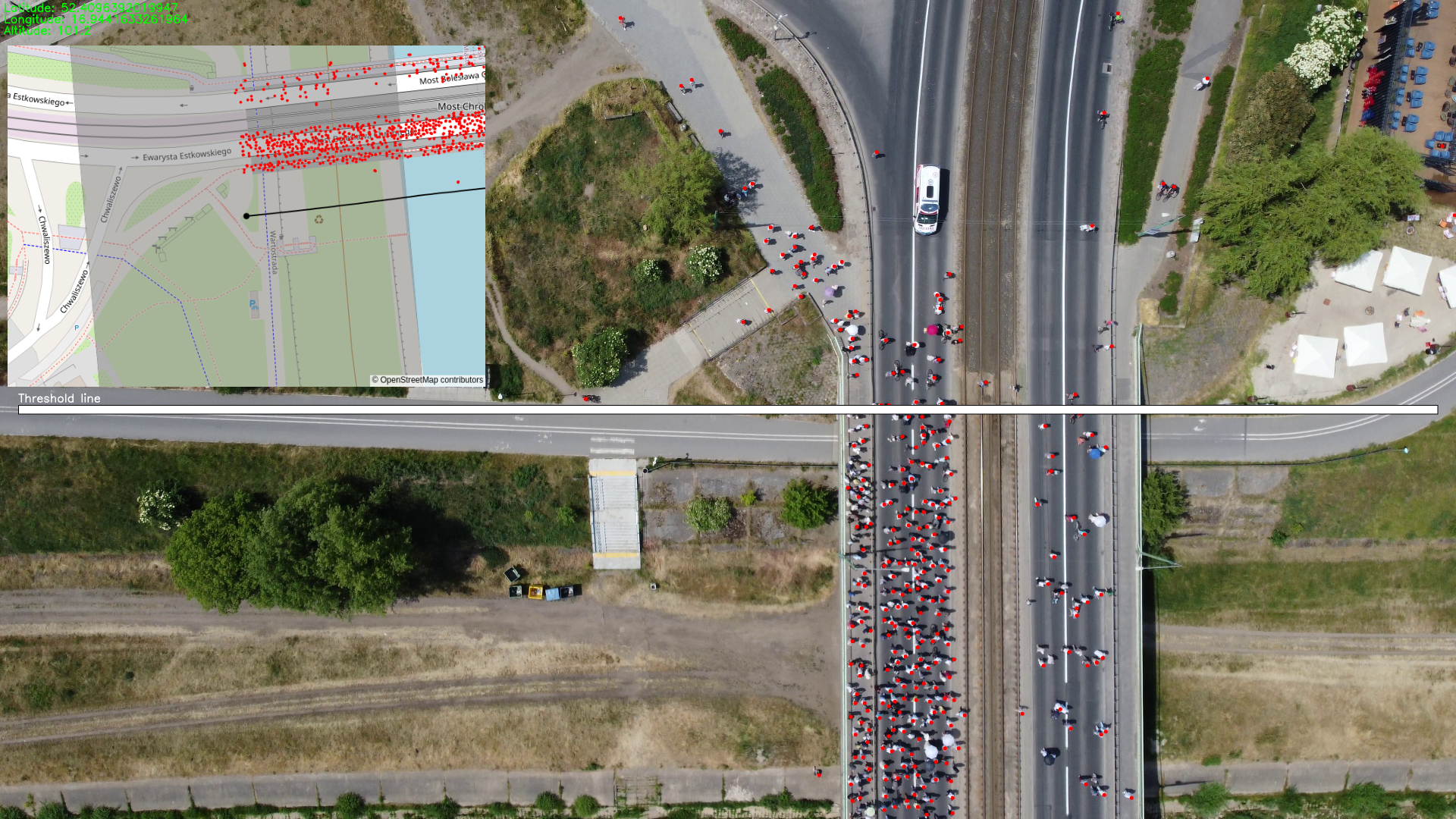}
         \caption{Frame 1280}
     \end{subfigure}

        \caption{Four example frames from a video recording of a march. Each frame contains people detections marked in red, the white threshold line indicating the consideration cut-off, and a mini-map displaying the global coordinates of the detected people.}
        \label{fig:peoplecouingapp}
\end{figure}

An application for processing drone videos was developed based on the proposed algorithm trained on the UP-COUNT dataset. This application detects people in video frames and integrates this information with drone sensor measurements to generate a map of objects. Using camera calibration parameters, the drone's relative altitude, and gimbal position, our system triangulates people's head locations and maps them accordingly. Due to the lack of object tracking capabilities, a horizontal threshold line at the image centre is used to count individuals only once (an object is added when it crosses the threshold line). Unfortunately, this approach requires the drone to move in the opposite direction of the crowd or in the same direction at a higher speed, introducing processing limitations. Sample frames from the processing of an example video are shown in~\Cref{fig:peoplecouingapp}. The red dots on the image indicate people's locations detected using the neural network. The figure also illustrates the white threshold line; when detected points cross this line, their global coordinates are calculated. Additionally, a mini-map in the top-left corner of the frame shows these global coordinates for visualisation purposes. As a result, the map with detected people coordinates is created, enabling further crowd flow processing and density analysis.

\section{Conclusions}

This paper introduces the Dot localisation method with the Pixel Distill module for the tiny objects localisation task using high-resolution UAV-recorded images. Extensive evaluation demonstrated that this method surpasses the performance of previous state-of-the-art methods on the DroneCrowd dataset and outperforms results obtained by contemporary approaches adapted from other remote sensing tasks. Additionally, alongside the proposed method, the UP-COUNT dataset has been shared, comprising ten thousand images annotated with point-oriented object labels dedicated to people's localisation. Noteworthy features of this dataset, such as a moving camera and variations in altitudes and locations, contribute to its high data diversity and suitability for novel applications, as demonstrated.

\subsection{Limitations}
The proposed Pixel Distill module is designed to decrease the image resolution, limiting to dimensions divisible by 32. Furthermore, applicable transformations can only reduce the dimensionality by factors of 2, 4, 8, etc. Because the module is mainly developed for high-resolution images, where the size reduction can lead to significant computational savings, the benefit of the module is expected to be smaller for lower-resolution images, and it might be less important for images with resolutions similar to the model input.

\subsection{Future work}
Moving forward, our next objective is to prepare tracking labels for the UP-COUNT's test sequences, facilitating the association of point-oriented objects and enabling tracking evaluation for better individual recognition. An equally exciting approach would be to leverage altitude information to address scale variation issues and research its influence on false positive and false negative predictions.

\section*{Ethic Statement}
The UP-COUNT dataset maintains data privacy by design. Thanks to the aerial perspective, which captures individuals as tiny figures, the dataset mitigates privacy concerns by minimising identifiable features.

While this research focuses on civilian applications such as crowd monitoring for public safety and crowd management, it is important to acknowledge the method could be potentially adapted for applications in surveillance and target identification, highlighting the importance of responsible development and usage.

\section*{Data Availability Statement}

The data supporting the findings of this research are publicly available at \url{https://doi.org/10.5281/zenodo.12683104}.



\end{document}